%% file: acl_latex.tex
\pdfoutput=1

\documentclass[11pt]{article}

\usepackage[final]{acl}

\usepackage{times}
\usepackage{latexsym}

\usepackage[T1]{fontenc}

\usepackage[utf8]{inputenc}

\usepackage{microtype}

\usepackage{inconsolata}

\usepackage{graphicx}
\usepackage{amsmath}
\usepackage{cleveref}
\usepackage{amssymb}
\usepackage{mathtools}
\usepackage{amsthm}
\RequirePackage{times}

\usepackage{fancyhdr}
\usepackage{algorithm}
\usepackage{natbib}
\usepackage{eso-pic} 
\usepackage{forloop}
\usepackage{url}

\theoremstyle{plain}

\theoremstyle{definition}

\theoremstyle{remark}

\newcommand{\ours}{\texorpdfstring{FLARE}\xspace}

\usepackage[textsize=tiny]{todonotes}
\usepackage{hyperref}
\usepackage{booktabs}
\usepackage{multirow}
\usepackage{url}
\usepackage[table,xcdraw]{xcolor}
\usepackage[normalem]{ulem}
\useunder{\uline}{\ul}{}
\usepackage{adjustbox}
\usepackage{wrapfig}
\usepackage{float}
\usepackage{dsfont}
\usepackage{wrapfig}
\usepackage{longtable}
\usepackage{tabularx}
\usepackage{listings}
\usepackage{caption}
\usepackage{algpseudocode}

\usepackage{amsmath}
\usepackage{enumitem}

\definecolor{darkblue}{rgb}{0.0, 0.0, 0.5}

\definecolor{darkblue}{rgb}{0.0, 0.0, 0.5}

\hypersetup{
    colorlinks=true,
    linkcolor=darkblue,      
    citecolor=darkblue,      
    urlcolor=darkblue        
}

\title{FLARE: Faithful Logic-Aided Reasoning and Exploration}

\newcommand\blfootnote[1]{%
  \begingroup
  \renewcommand\thefootnote{}\footnote{#1}%
  \addtocounter{footnote}{-1}%
  \endgroup
}

\author{
Erik Arakelyan$^{\dagger15}$ \quad Pasquale Minervini$^{2}$$^{3}$ \quad Pat Verga$^4$ \quad Patrick Lewis$^4$ \quad Isabelle Augenstein$^1$ \\
$^1$University of Copenhagen \qquad $^2$University of Edinburgh \\ $^3$Miniml.AI \qquad $^4$Cohere \qquad $^5$NVIDIA \\
\texttt{earakelyan@nvidia.com} \qquad \texttt{augenstein@di.ku.dk} \qquad  \texttt{p.minervini@ed.ac.uk} \\ \texttt{\{pat, patrick\}@cohere.com}\\
}

\begin{document}

\maketitle
\blfootnote{$^{\dagger}$Corresponding author.}

\begin{abstract}
%
%
%
%
%
Modern Question Answering (QA) and Reasoning approaches with Large Language Models (LLMs) commonly use Chain-of-Thought (CoT) prompting but struggle with generating outputs faithful to their intermediate reasoning chains.
While neuro-symbolic methods like Faithful CoT (F-CoT) offer higher faithfulness through external solvers, they require code-specialized models and struggle with ambiguous tasks.
%
%
%
We introduce \textbf{F}aithful \textbf{L}ogic-\textbf{A}ided \textbf{R}easoning and \textbf{E}xploration (\textbf{\ours}), which uses LLMs to plan solutions, formalize queries into logic programs, and simulate code execution through multi-hop search without external solvers.
%
%
%
Our method achieves SOTA results on $\mathbf{7}$ out of $\mathbf{9}$ diverse reasoning benchmarks and $3$ out of $3$ logic inference benchmarks while enabling measurement of reasoning faithfulness.
%
%
%
We demonstrate that model faithfulness correlates with performance and that successful reasoning traces show an $18.1\%$ increase in unique emergent facts, $8.6\%$ higher overlap between code-defined and execution-trace relations, and $3.6\%$ reduction in unused relations.
%
%
\end{abstract}

\section{Introduction} \label{sec:intro}
\begin{figure*}[!t]
    \centering
    \includegraphics[width=\textwidth, clip]{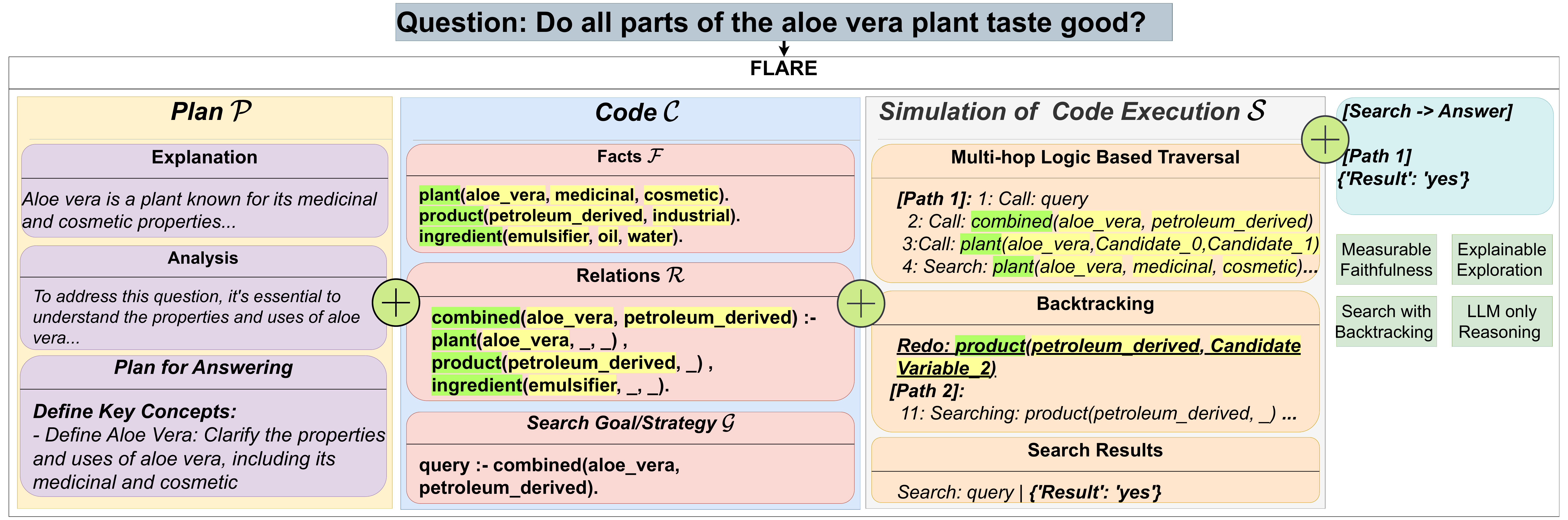}
    \caption{A depiction of the \emph{plan}, \emph{code} and simulated \emph{search} in {\ours}. Each module is generated separately and iteratively, allowing us to obtain the final answer. The green and yellow highlighted text shows the overlap between the facts and the relations between the code and the simulated search.}
    \label{fig:flare}
\end{figure*}

%
%
Complex Reasoning in natural Question Answering (QA) tasks requires exploring a problem space with formalized facts, relations, commonsense knowledge and logical implications. 
In line with this, LLMs have been enhanced with CoT \citep{wei2022chain} prompting, which supplements the QA process by generating intermediate reasoning chains given a set of in-context examples \citep{NEURIPS2020_1457c0d6}, as shown in \cref{fig:flare}.
This allowed for advancement in commonsense \citep{madaan2022language}, symbolic \citep{wang2022self,sprague2024cot} and mathematical \citep{jie2023design} reasoning.
Although CoT allows for a problem exploration in natural language steps, such an approach has been shown to cause performance degradation for reasoning tasks involving multi-step planning~\citep{valmeekam2022large,suzgun-etal-2023-challenging}, problem exploration~\citep{yao2022react}, and arithmetic tasks~\citep{hendrycks2021measuring, madaan2022text}.
These discrepancies arise as CoT suffers from a limited ability to decompose, search, verify and backtrack using intermediate rationale chains~\citep{yao2022react}, cascading hallucinations and errors \citep{DBLP:conf/nips/LingFLHLMS23} and that natural language might not be an optimal representation for describing the reasoning process~\citep{DBLP:conf/icml/0002LZCHSL0XI24}.
Simultaneously, LLM output has been shown to be unfaithful and inconsistent w.r.t. the intermediate CoT rationale \citep{DBLP:conf/acl/JacoviBBHHT0AG24,lanham2023measuring,DBLP:conf/nips/TurpinMPB23}.
%

%
%
%
To mitigate the problem of CoT faithfulness and allow for more robust reasoning during QA, \citet[Faithful CoT]{DBLP:conf/ijcnlp/LyuHSZRWAC23} and Logic-LM~\citep{DBLP:conf/emnlp/PanAWW23} suggested generating code which is further executed using an external symbolic solver.
Producing and executing code enables the generation of outputs guided by external solvers, leveraging search with backtracking to explore the problem space effectively.
However, strict translations of natural language queries into code, such as \emph{autoformalisation}~\citep{DBLP:conf/mkm/Szegedy20,DBLP:conf/mkm/WangKU18}, is a non-trivial task involving direct inference of implicit commonsense and domain-specific knowledge and the ability to align abstract and informal concepts directly to constrained formal definitions for further execution~\citep{DBLP:conf/nips/WuJLRSJS22}.
An example query, \emph{``Do all parts of the aloe vera plant taste good?''}, is challenging to formalize or address with a strict algorithmic solution, as it requires interpretative, deductive and context-dependent reasoning, referred to as soft or fuzzy reasoning.
Using external solvers makes such fuzzy reasoning impossible and requires consistently generating syntactically correct executable code.
While some LLMs have coding capabilities stemming from their pretraining \citep{DBLP:journals/corr/abs-2406-00515,DBLP:journals/corr/abs-2408-10914}, relative code consistency is more probable with models explicitly trained for coding \citep{DBLP:journals/corr/abs-2107-03374}.
To overcome these problems, we propose Faithful Logic-Aided Reasoning and Exploration (\ours), an interpretable method that allows for planning, fuzzy reasoning, and traversing the problem space with backtracking, exact task decomposition, and measuring faithfulness.
In \ours, given a natural language query, we prompt an LLM to sequentially generate a \emph{plan} that includes an analysis and the logical steps necessary for formalising and answering the question, a logic programming~\citep{wielemaker2012swi} \emph{code} that allows formalising the query into a set of facts, relations and their composition forming the space for exploring that query and the \emph{search}, which is an LLM-generated code execution simulation.
An illustration of {\ours} can be seen in \cref{fig:flare}.
This work focuses on models that \textbf{have not} been explicitly trained on CoT on other reasoning traces, as these models have been shown to struggle with generalisation towards differing reasoning paradigms \citep{chen2024not}, consistency in intermediate reasoning steps \citep{wang2025thoughts} and instruction following \citep{zhang2025s1}.
In our framework, the generated code must not be consistently executable by an external solver, allowing for the soft-formalisation of natural language. Although we see that even generalist LLMs are able to produce executable code in $\geq 50\%$ of cases.
{\ours} allows us to measure the faithfulness of the outcome w.r.t. the simulated code execution by directly comparing the search paths produced by the external solver to that LLM generation.
This comparison also allows for pinpointing model hallucinations and inconsistencies.
We systematically study the effectiveness of our method using $4$ general-purpose LLMs of varying scales across $9$ diverse QA and $3$ logical inference benchmarks, covering Math World Problems, Multi-hop QA, Relation inference, deductive and analytical reasoning and show that our method achieves state-of-the-art results in $7$ out of $9$ QA datasets and $2$ out of $3$ logic datasets in comparison to CoT, F-CoT and Logic-LM.
We also show that the method is competitive for models tuned for coding, with an average overall increase of $16\%$ over F-Cot and $10\%$ over CoT.
%
%
%
Our key contributions are the following: \textbf{(i)} We introduce {\ours}, a novel paradigm for logic-aided and interpretable formalisation and search over the problem space in QA and logic reasoning tasks. \textbf{(ii)} We perform a systematic evaluation across $9$ QA and $3$ logical inference benchmarks and $4$ models of varying scales, showing the advantages of using {\ours} for QA in a few-shot setup over prior approaches. \textbf{(iii)} The modularity of {\ours} allows defining a simple ingrained method for measuring model faithfulness, which is further shown to be strongly correlated with performance. \textbf{(iv)} We further show that using {\ours} allows us to interpretably and rigorously detect hallucinations along with sub-optimal and inconsistent reasoning patterns.


\section{Related Work}
\label{sec:related}

\paragraph{Reasoning in Natural Language}
\paragraph{Reasoning in Natural Language}

Few-shot prompting~\citep{DBLP:conf/nips/BrownMRSKDNSSAA20} improves LLM reasoning, and extensions like Chain-of-Thought (CoT)\citep{wei2022chain}, “think step by step’’\citep{DBLP:conf/nips/KojimaGRMI22}, and Least-to-Most~\citep{DBLP:conf/iclr/ZhouSHWS0SCBLC23} explicitly decompose problems into intermediate steps.
Despite their promise, these methods exhibit arithmetic errors~\citep{DBLP:conf/nips/LewkowyczADDMRS22,DBLP:conf/nips/HendrycksBKABTS21} and logical inconsistencies~\citep{DBLP:journals/corr/abs-2209-07686}.
Planning-based variants introduce a separate plan–execute loop~\citep{DBLP:conf/iclr/YaoZYDSN023,DBLP:conf/acl/WangXLHLLL23}.
The \emph{plan} stage in {\ours} draws on these ideas but focuses on generating a natural-language strategy for later formalisation into code.
\paragraph{Reasoning with Search}

Recent work augments LLM reasoning by explicitly searching the problem space.
Self-consistency decoding~\citep{DBLP:conf/iclr/0002WSLCNCZ23} samples multiple chains of thought and selects the majority answer, while Tree-of-Thoughts~\citep[ToT;][]{DBLP:conf/nips/YaoYZS00N23} performs tree-structured exploration with LLM-evaluated states.
Later methods adapt classical search—DFS, BFS~\citep{DBLP:conf/aaai/BestaBKGPGGLNNH24}, A$^*$\citep{DBLP:journals/corr/abs-2402-14083}, and hybrids\citep{DBLP:journals/corr/abs-2404-03683}—via direct tuning, imitation learning~\citep{DBLP:conf/nips/YangSAN22}, or few-shot prompting~\citep{DBLP:journals/corr/abs-2404-01230}.
So far, evaluations focus on toy puzzle and algorithmic domains such as the 24 Game, Countdown, Sorting, mazes, and Sokoban~\citep{DBLP:conf/nips/YangSAN22,wiki:Countdown,DBLP:conf/aaai/BestaBKGPGGLNNH24,DBLP:journals/corr/abs-2402-14083}.
Although the \emph{search} module of {\ours} shares this multi-path exploration spirit, it targets more general tasks and yields interpretable multi-hop reasoning via simulated code execution.
\paragraph{Reasoning with Formalisation}
Another research direction explores formalising natural language queries into code~\citep{DBLP:conf/icml/GaoMZ00YCN23,DBLP:conf/icml/0002LZCHSL0XI24} or pseudo-code~\citep{DBLP:journals/corr/abs-2404-02575,DBLP:journals/corr/abs-2404-03683}.
This enables translating queries into strict structures, delegating reasoning and search to deterministic solvers such as Python~\cite{DBLP:journals/tmlr/ChenM0C23}, PDDL~\cite{DBLP:conf/ijcnlp/LyuHSZRWAC23, DBLP:journals/corr/abs-2304-11477}, or DataLog~\cite{DBLP:conf/ijcnlp/LyuHSZRWAC23}.
Models can synthesize programs~\citep{DBLP:journals/corr/abs-2108-07732, DBLP:conf/iclr/NijkampPHTWZSX23} and benefit from code in numerical and algorithmic reasoning~\citep{DBLP:journals/tmlr/ChenM0C23,DBLP:conf/icml/GaoMZ00YCN23}, yet their use for general QA remains underexplored.
This is due to the challenge of translating natural language into strictly executable code~\citep{DBLP:conf/nips/WuJLRSJS22}, the syntactic rigidity of underrepresented programming languages during pre-training~\citep{DBLP:journals/corr/abs-2404-00971}, and the need for models explicitly tuned for coding~\citep{DBLP:journals/corr/abs-2107-03374}.
Additionally, relying on external solvers restricts soft reasoning over commonsense knowledge and implications. In {\ours}, we formalise queries as logic programs in Prolog during the \emph{code} generation step but do not require executability or external solvers at inference. This allows LLMs to simulate code execution via soft reasoning over logic-based traversals—similar to Prolog—while avoiding the need for code-specific tuning.
%
\input{results}

\paragraph{Reasoning Faithfulness}

An explanation is considered \emph{faithful} if it explicitly and accurately describes the reasoning process of the model during inference \citep{DBLP:conf/dsaa/GilpinBYBSK18, DBLP:conf/acl/JacoviG20}. In the context of prompting techniques such as CoT, we are interested in the faithfulness of the intermediate reasoning chains towards the final output. Faithful intermediate reasoning chains should not just look \emph{plausible} \citep{herman2017promise} but have exact reflections of the problem exploration and reasoning used to arrive at the final answer. Natural language reasoning chains prevalent in CoT and similar methods are shown to be unfaithful, either masking the reasoning biases \citep{DBLP:conf/nips/TurpinMPB23} of the model or outright ignoring the intermediate reasoning \citep{DBLP:journals/corr/abs-2307-13702}. In {\ours}, we introduce a method to seamlessly measure the faithfulness of the final outcome w.r.t. completed search.

\section{Methodology}
\label{sec:method}
\subsection{LLM-Simulated Search}
\label{subsec:generate}

{\ours} comprises three modules for generating a \emph{plan}, \emph{code} and simulated \emph{search} for answering a natural language query $\mathcal{Q} = \{T^{\mathcal{Q}}_1 \dots T^{\mathcal{Q}}_{|\mathcal{Q}|}\}$, where each $T^{\mathcal{Q}}_i$ is a token in the query $\mathcal{Q}$.

\paragraph{Generating A Plan}
For each query $\mathcal{Q}$, given an LLM $\mathcal{M}$, we initially use instructions $\mathcal{I}^{\mathcal{P}}$ to prompt it to generate a \emph{plan} $\mathcal{P}$, which should be comprised of task explanation, analysis and a plan for further formalising the query. An example of this can be seen in the \emph{plan} section in \cref{fig:flare}. We use in-context few shot examples $\mathcal{E}_{\mathcal{P}}$ of such \emph{plan} generations 
for obtaining the final plan:
\begin{align}
    \mathcal{P}_i \sim  p_\mathcal{M}(T^{\mathcal{P}}_i \mid T^{\mathcal{P}}_{:i-1},\mathcal{E}_{\mathcal{P}}, \mathcal{Q},\mathcal{I}^{\mathcal{P}}),
\end{align}
\noindent where $\mathcal{P}_i\text{ and } T^{\mathcal{P}}_i$ is the $i$-th token in the generated \emph{plan} $\mathcal{P}$ and $p_\mathcal{M}$ is the probability of the next token over the vocabulary obtained from model $\mathcal{M}$.

\paragraph{Generating Code}

After generating the \emph{plan}, we use instructions $\mathcal{I}^{\mathcal{C}}$ to prompt the LLM $\mathcal{M}$ to generate a Prolog code $\mathcal{C}$, an example of which can be seen in \cref{fig:flare}. We append executable code generation samples $\mathcal{C}_{\text{sample}}$ to the previous in-context examples $\mathcal{E}_{\mathcal{P}}$ and obtain few-shot code generation demonstrations $\mathcal{E}_{\mathcal{C}} =[\mathcal{E}_{\mathcal{P}};\mathcal{C}_{\text{sample}}]$
\begin{gather}
\label{eq:code}
     \mathcal{C}_i \sim  p_\mathcal{M}(T^{\mathcal{C}}_i \mid T^{\mathcal{C}}_{:i-1} \mathcal{E}_{\mathcal{C}}, \mathcal{Q}, \mathcal{I}^{\mathcal{P}},\mathcal{P},\mathcal{I}^{\mathcal{C}})
     \\
     \mathcal{F}_{\text{\emph{code}}}, \mathcal{R}_{\text{\emph{code}}} , \mathcal{G}_{\text{\emph{code}}} = \text{EXTRACT}(\mathcal{C}_i), \nonumber
 \end{gather}
\noindent where $\mathcal{C}_i \text{ and }T^{\mathcal{C}}_i$ is the $i$-th token in the generated \emph{code} $\mathcal{C}$. We detail the benefits of Prolog and the reasoning behind our choice in \cref{append:prolog}.

\begin{figure*}[t!]
    \centering
    \includegraphics[clip=true,width=\textwidth]{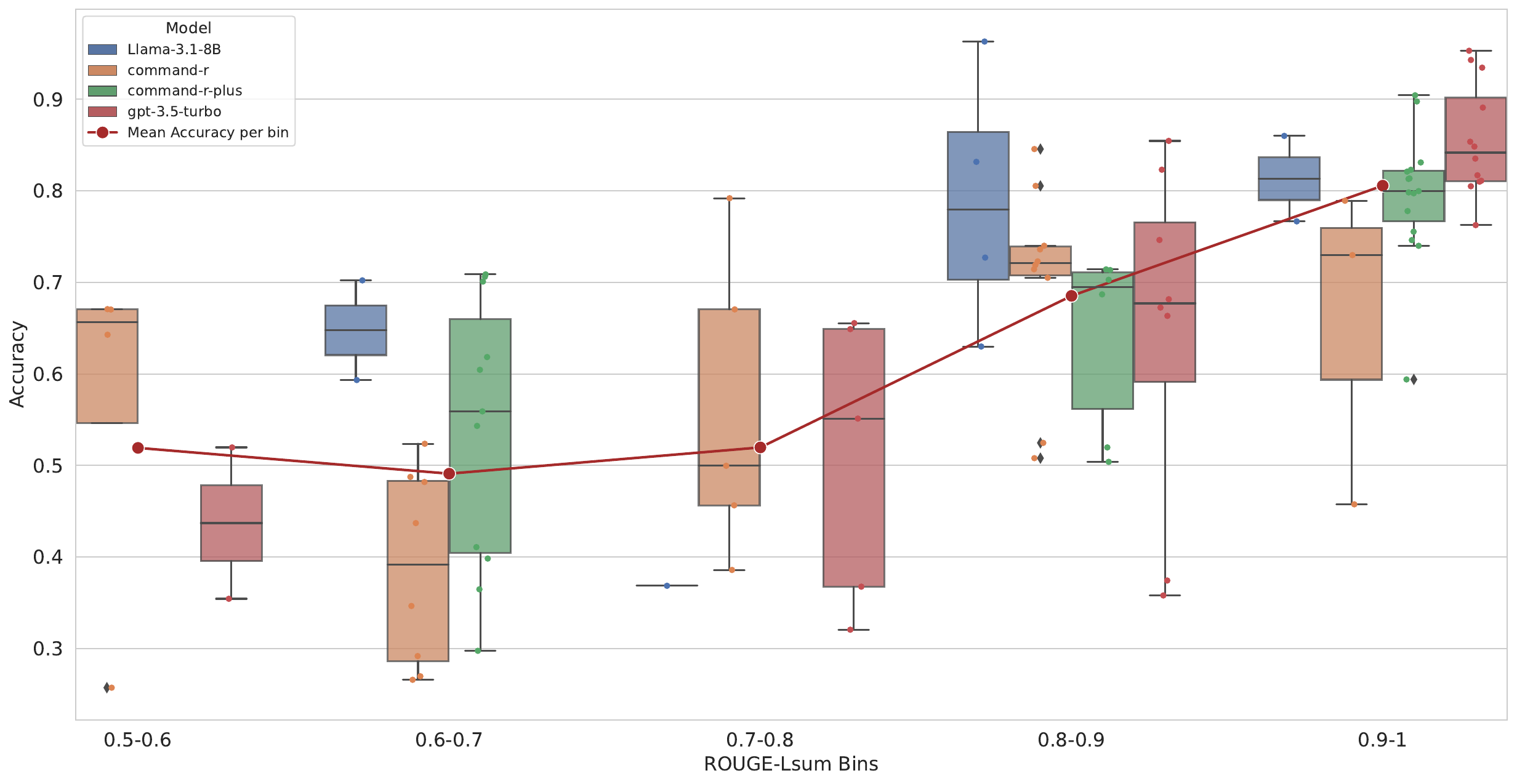}
    \caption{The trend of mean model accuracy w.r.t mean faithfulness for all the models.}
    \label{fig:faith_vs_acc}
\end{figure*}

\input{logiclm_compare}

\paragraph{Simulating Search}

After generating the logic-programming \emph{code}, we want to simulate program execution by generating a problem space traversal trace with our LLM $\mathcal{M}$. We use instructions $\mathcal{I}^{\mathcal{S}}$ and update our in-context samples by appending search traces $\mathcal{S}_{\text{sample}}$ constructed from Prolog execution of sample codes $\mathcal{C}_{\text{sample}}$, i.e. $\mathcal{E}_{\mathcal{S}} =[\mathcal{E}_{\mathcal{C}};\mathcal{S}_{\text{sample}}]$:
\begin{gather}
\label{eq:search}
     \mathcal{S}_i \sim  p_\mathcal{M}(T^{\mathcal{S}}_i \mid T^{\mathcal{S}}_{:i-1} \mathcal{E}_{\mathcal{C}}, \mathcal{Q}, \mathcal{I}^{\mathcal{P}},\mathcal{P},\mathcal{I}^{\mathcal{C}},\mathcal{C},\mathcal{I}^{\mathcal{S}}) \\
     \mathcal{A}_{\text{\emph{search}}},
     \mathcal{F}_{\text{\emph{search}}}, \mathcal{R}_{\text{\emph{search}}} = \text{EXTRACT}(\mathcal{S}_i), \nonumber
 \end{gather}
\noindent where $T^{\mathcal{S}}_i$ is the $i$-th token in the generated \emph{search} trace $\mathcal{S}$. During iterative problem space traversal, we can segment the facts $\mathcal{F}_{\text{\emph{search}}}$, relations $\mathcal{R}_{\text{\emph{search}}}$, completed and backtracked paths with their answers $\mathcal{A}_{\text{\emph{search}}}$ used during the search simulation. To get the final answer we update in-context samples with their correct final answers $\mathcal{A}_{\text{sample}}$ from the executed search $\mathcal{S}_{\text{sample}}$, $\mathcal{E}_{\mathcal{A}} =[\mathcal{E}_{\mathcal{S}};\mathcal{A}_{\text{sample}}]$ and use instructions $\mathcal{I}^\mathcal{A}$ to obtain the final answer from the model.
 \begin{align}
      \mathcal{A}_{\text{\emph{Final}}} \sim  p_\mathcal{M}(T^{\mathcal{A}}_i \mid T^{\mathcal{A}}_{:i-1} \mathcal{E}_{\mathcal{C}}, \mathcal{Q}, \\\mathcal{I}^{\mathcal{P}},\mathcal{P},\mathcal{I}^{\mathcal{C}},\mathcal{C},\mathcal{I}^{\mathcal{S}},\mathcal{S},\mathcal{I}^\mathcal{A}) \nonumber
 \end{align}
The prompts used for generating each part in {\ours} can be seen in \cref{append:prompts} along with a complete example in \cref{tab:flare_ex_complete} and a pseudo-code in \cref{algo:flare}.     

\subsection{Detecting Reasoning Inconsistencies}
\label{subsec:reasoning_inc}

For each query $\mathcal{Q}$ given the \emph{code} $\mathcal{C}$ and the simulated \emph{search} $\mathcal{S}$ along with the extracted facts $\mathcal{F}_{\text{\emph{code}}}, \mathcal{F}_{\text{\emph{search}}}$ and relations $\mathcal{R}_{\text{\emph{code}}}, \mathcal{R}_{\text{\emph{search}}}$ from each designated module, we aim to detect the inconsistencies during the reasoning process of the LLM. We use exact string matching between all these facts and relations in \emph{code} and simulated \emph{search}.
\begin{gather}
\forall i, \exists j \quad \text{such that} \quad \mathcal{F}_{\text{code}}^i = \mathcal{F}_{\text{search}}^j \quad \\ \text{and} \quad \forall v, \exists q \quad \mathcal{R}_{\text{code}}^v = \mathcal{R}_{\text{search}}^q \nonumber \\
\forall j, \exists i \quad \text{such that} \quad \mathcal{F}_{\text{code}}^i = \mathcal{F}_{\text{search}}^j \quad \\ \text{and} \quad \forall q, \exists v \quad \mathcal{R}_{\text{code}}^v = \mathcal{R}_{\text{search}}^q \nonumber
\end{gather}
With this framework in mind, we define two reasoning failure modes. 
In the \emph{first} failure mode, given that some fact or relation was used in the simulated \emph{search} but did not exist in the generated \emph{code}, i.e. $\exists j \text{ such that } \mathcal{F}_{\text{search}}^j \notin \mathcal{F}_{\text{code}}$, we claim that the LLM has \emph{hallucinated}. We postulate that the model either produced incomplete knowledge during formalisation to \emph{code} or created a piece of non-existing information during the \emph{search}. We do not consider facts that emerged during a direct inference step within the simulated search during our calculation. For example, if we are dealing with a mathematical query $4\cdot (5+6) = ?$, the search would involve separately evaluating the expression $5+6=11$. In this case, $11$ will not be treated as a hallucinated fact within the search but rather as an emergent fact obtained from direct inference.   
The \emph{second} failure mode is the reciprocal case, where a fact or relation present in the \emph{code} is not used during the \emph{search}. We refer to this phenomenon as \emph{sub-optimal reasoning} as it shows that the LLM could not explore the problem space completely or injected unsuitable knowledge during formalisation into \emph{code}.

\subsection{Measuring Faithfulness}

We propose a method to measure the faithfulness of the LLM reasoning process when using {\ours}. As mentioned in \cref{subsec:generate}, for each query in a dataset $\mathcal{D} = [\mathcal{Q}_1, \dots ,\mathcal{Q}_{|\mathcal{D}|}]$, we generate a set of codes $\Phi = [\mathcal{C}_1, \dots ,\mathcal{C}_{|\Phi|}]$ and simulated problem space searches $\Psi = [\mathcal{S}_1, \dots, \mathcal{S}_{|\Psi|}]$. We use the Prolog engine to execute all of the codes $\Phi$ and obtain a set of correctly written programs $\Phi^\prime$ and exact search paths $\Psi^\prime$. As we do not require explicit programmatic correctness during inference in {\ours} for any code $\mathcal{C}_i$, some Prolog executions resulting in an error are filtered out in $\Psi^\prime$. To assess model reasoning faithfulness towards code formalisations, we compare the search paths $\Phi^\prime$ obtained from Prolog execution with their designated counterparts $\Phi_{\text{\emph{gen}}}^\prime$ generated by the LLM from the same code. We use ROUGE \citep{lin-2004-rouge} to compute the matching score for each executed and simulated search path. In particular, we use ROUGE-Lsum, which uses the longest common subsequence (LCS) over each line to obtain the final score. This method fits our cause as a line in a Prolog search execution represents a single logic step within the traversal. This allows us to measure the similarity of the reasoning contents and structure in exact and simulated searches. We have also used other string-matching techniques, all of which show the same trends; thus, we report our results with ROUGE-Lsum.

\input{prompt_only}
\section{Experimental Setup}
\label{sec:ex_setup}

\paragraph{Datasets}
To evaluate {\ours}, we use a benchmark of $9$ tasks spanning Math Word Problems (MWP), multi-hop QA, relation inference, and $3$ logical reasoning datasets. For numerical and mathematical reasoning, we follow CoT~\citep{wei2022chain} and include GSM8K~\citep{DBLP:journals/corr/abs-2110-14168}, SVAMP~\citep{DBLP:conf/naacl/PatelBG21}, MultiArith~\citep{DBLP:conf/emnlp/RoyR15}, ASDiv~\citep{DBLP:conf/acl/MiaoLS20}, and AQuA~\citep{DBLP:conf/acl/LingYDB17}. GSM8K, SVAMP, MultiArith, and ASDiv focus on elementary and middle school arithmetic with integer or decimal answers. AQuA involves multiple-choice symbolic reasoning with expressions not explicitly defined in the query.
We also test {\ours} on three multi-hop QA tasks. StrategyQA~\citep{DBLP:journals/tacl/GevaKSKRB21} requires boolean reasoning with sub-goal decomposition (e.g., \emph{``Do all parts of the aloe vera plant taste good?''} in \cref{fig:flare}). We further use the Date and Sports Understanding subsets from BIG-Bench~\citep{srivastava2023beyond}, which involve temporal and feasibility-based reasoning.
For relation inference, we use CLUTRR~\citep{DBLP:conf/emnlp/SinhaSDPH19}, which requires deducing familial relations from partial graph descriptions in natural language.
We evaluate logical reasoning using ProntoQA~\citep{DBLP:conf/iclr/Saparov023}, AR-LSAT~\citep{DBLP:journals/corr/abs-2104-06598}, and LogicalDeductions from BIG-Bench~\citep{DBLP:journals/tmlr/SrivastavaRRSAF23}, focusing on the challenging subsets of~\citep{DBLP:conf/emnlp/PanAWW23}. These cover deductive, analytical, and logical tasks. Dataset details and examples are in \cref{tab:data_stat} of \cref{append:prompts}. We also study how model size affects performance and faithfulness (\cref{sec:scale_effect}).

\paragraph{Benchmarks}
We compare {\ours} with CoT~\citep{wei2022chain}, which uses natural language reasoning chains, and with F-CoT~\citep{DBLP:conf/ijcnlp/LyuHSZRWAC23} and Logic-LM~\citep{DBLP:conf/emnlp/PanAWW23}, which formalise queries into code and delegate reasoning to external solvers. Evaluated models include Llama3.1 (8B)\citep{DBLP:journals/corr/abs-2407-21783}, CmDR (30B) and CmDR+ (100B)\citep{cohere2024commandr}, and GPT-3.5~\citep{DBLP:conf/nips/BrownMRSKDNSSAA20} ($\geq$100B~\citep{DBLP:journals/corr/abs-2303-10420}). As OpenAI Codex (code-DaVinci-002)~\citep{DBLP:journals/corr/abs-2107-03374} used in F-CoT has been deprecated, we replace it with the new GPT3.5 as suggested by OpenAI and recalculate the results.

\section{Results}
\label{sec:results}

\subsection{Few-shot prompting}

To evaluate {\ours}, we use a set of models of varying sizes on diverse benchmarks, as defined in \cref{sec:ex_setup}. We compare the performance of each model while using {\ours}, CoT and F-CoT prompting. The results for F-CoT and CoT on all the models are computed using the codebase of the original study \citep{DBLP:conf/ijcnlp/LyuHSZRWAC23}. We additionally compare Logic-LM and {\ours} using the logic reasoning benchmarks proposed in \cite{DBLP:conf/emnlp/PanAWW23}.

\paragraph{LLMs for general reasoning} 

Our results, presented in \cref{tab:results_main}, show that using {\ours} allows the LLMs to achieve state-of-the-art results on $7$ out of $9$ datasets, with an average $28\%$ increase over CoT. We can see a clear trend that {\ours} increases the performance compared to CoT and F-CoT for all the models of varying scales. We also see that LLMs not explicitly tuned for coding suffer massive degeneracies when using F-CoT. We postulate that they cannot consistently produce executable programs that satisfy a predefined scheme in F-CoT, thus resulting in an error during execution. This further highlights the value of simulating program execution using an LLM instead of external solvers.
The results show that using {\ours} yields more benefit on datasets that require longer chains of multi-hop and symbolic reasoning, like AQuA and StrategyQA.
Our findings in \cref{tab:logiclm_res} show that \ours achieves state-of-the-art results on $2$ out of $3$ logic inference benchmarks with $10\%$ increase over CoT and $~7\%$ increase over Logic-LM. Following the practice in \citep[Logic-LM]{DBLP:conf/emnlp/PanAWW23}, we also add $2$ iterations of code self-refinement to {\ours} and show that the model model is able to achieve SOTA results on all $3$ benchmarks.

\paragraph{LLMs for code generation}

To understand the effect of {\ours} on models tuned for coding, we use GPT3.5 \citep{NEURIPS2020_1457c0d6} as it was the OpenAI suggested succession model for Codex \citep{DBLP:journals/corr/abs-2107-03374} which is used in F-CoT and possesses strong coding capabilities \citep{DBLP:journals/corr/abs-2303-10420}. The results in \cref{tab:results_main} show that using {\ours} is beneficial for models tuned for coding and boosts accuracy with a $16\%$ increase over F-CoT and $9\%$ over CoT. The reason is that many natural language queries with non-trivial formalisations are more suited to be tackled with more commonsense soft reasoning than direct code execution. This is evident in \cref{tab:results_main} where {\ours} and CoT are consistently better than F-CoT in StrategyQA, Sports and CLUTRR. 
The opposite case of numeric and algorithmic heavy reasoning tasks is also covered by {\ours} as it maintains strong performance similar to F-CoT on MWP problems \cref{tab:results_main}.
Consequently, {\ours} allows combining algorithmic formalisation with simulated soft-reasoning, circumventing the pitfalls of using a deterministic external solver while still producing a query formalisation and problem space traversal. 

\subsection{Is simulating search useful?}
\input{reasoning_stats_basic}

To understand if simulating a search over the problem space is useful, we compare the performance of {\ours} where we only generate the \emph{plan} without the subsequent \emph{code} or \emph{search} components. We refer to this framework setup as \emph{plan-only}, which can be seen in \cref{fig:flare} if we were to use only the \emph{plan} for answer generation. We completed this ablation using CmDR, CmDR+, and GPT-3.5, and we used GSM8K, AQuA, and StrategyQA as our baselines. The results in \cref{tab:plan_only} confirm that all of the models suffer massive performance degradation from $61.1 \rightarrow 49.9$ when omitting the \emph{code} and the \emph{search} components of {\ours}. We hypothesise that this is caused by insufficient problem space exploration when using the \emph{plan-only} setting.   
Furthermore, we have already seen in \cref{tab:results_main} that in methods, like F-CoT, that do not use simulated problem space exploration for soft-reasoning and only rely on \emph{plan} and \emph{code}, the performance also deteriorates even resulting in a complete breakdown of reasoning over the designated datasets. This can be viewed as a constrained version of {\ours} with \emph{code-only} execution.
Consequently, our results show that simulating problem space traversal is highly beneficial as it avoids the pitfalls posed by \emph{plan-only} and \emph{code-only} modes by exploring the problem space more rigorously and soft-reasoning during that traversal instead of using external solvers.

\subsection{Faithful Reasoning Improves Performance}
\input{reasoning_stats_deep}

As described in \cref{sec:method}, using {\ours} allows us to measure the faithfulness of the LLM reasoning process by comparing the simulated problem space traversals ${\Phi}^\prime_\text{\emph{gen}}$ with actual traces ${\Phi}^\prime$ produced from a symbolic Prolog solver. To do this, we initially compute the percentage of syntactically correct executable code each LLM produces. We have observed that all of the models are capable of producing correct executable Prolog code in $67\%$ of cases on average and $\geq 50\%$ of cases at the very least. The complete details can be seen in the top part of \cref{fig:code_perc} in \cref{append:prolog}. This shows that the simulated searches ${\Phi}^\prime_\text{\emph{gen}}$ can be considered a representative sample that will be further used to accurately measure the faithfulness of the simulated search w.r.t. the generated code. After measuring the reasoning faithfulness for each model, we want to understand what impact it has on the performance of the LLM. In \cref{fig:faith_vs_acc}, we segment the models w.r.t. their ROUGE-Lsum scores. The results show that model performance is strongly positively correlated with reasoning faithfulness. 
However, we also observe that executing semantically precise code results in an accurate answer only in $47\%$ of cases on average. Refer to the bottom part of \cref{fig:code_perc} in \cref{append:prolog} for more details. Indeed, having a simulated search trace with a ROUGE-Lsum faithfulness score of $1$, would be equivalent to simply executing the program as proposed in F-CoT. Yet we have priorly shown that F-CoT struggles with reasoning tasks that are hard to formalise and require multi-hop commonsense and soft reasoning.
These two discoveries show that optimal LLM reasoning, conditioned on a search in the problem space, should be increasingly faithful toward the facts, relations and the search strategy defined within the code while simultaneously maintaining the capability for soft-reasoning along more abstractly defined concepts. Our results show that {\ours} allows LLMs to maintain a similar reasoning capacity.


\subsection{What is important during the search?}

We also analyze the reasons which can lead to optimal reasoning within an LLM. We calculate several statistics, like the average number of explored paths and the average and total hops and failures per path, for each model during the simulated traversal. The failure in a path is an invalidation of a solution for a sub-goal explored during the search, which is used for backtracking, as explained in \cref{sec:method}. Calculating these statistics is simple as the \emph{search} component of {\ours}, seen in \cref{fig:flare}, is a structured simulation of a Prolog trace, where each line contains a hop of reasoning inference.
We split these statistics for the reasoning paths that lead to correct or incorrect outcomes. Our results in \cref{tab:reasoning_stats} show that LLM performance and reasoning optimality are not directly connected to the amount of explored paths or multi-hop inferences per path. We also see that traces that lead to incorrect answers have a higher number of failures per path and in total. We hypothesise that LLMs with traces that were optimal for reasoning and led to correct answers could skip exploring degenerate solutions due to strong commonsense reasoning capabilities.
Further analyses focus on identifying inconsistencies and failure modes (\cref{subsec:reasoning_inc}). By comparing relations in code with those in search traces, we measure emergent hallucinations and unused relations, highlighting areas of sub-optimal reasoning. We also assess the uniqueness of emergent facts per inference hop, indicating the extent of problem-space exploration (\cref{tab:reasoning_stats_deep}).
The results in \cref{tab:reasoning_stats_deep} show consistently over each model that, on average, traces that lead to correct answers had a higher percentage of unique emergent facts (UEF) and overlap in the relations (OR) used between the code and search, while the portion of underutilized relations (UR) was lower. This means that optimal reasoning with an LLM requires a great degree of problem-space exploration with fewer relation hallucinations during the search and more relation utilization from the defined code. This aligns with our prior discoveries, which show a strong correlation between simulated search faithfulness towards the formalised code and model performance. 

\section{Conclusion}

This work introduces {\ours}, a novel approach for logic-aided interpretable formalisation and reasoning with simulated search over the problem space. We show that models of varying scales obtain state-of-the-art results compared to prompting paradigms like CoT, F-CoT and Logic-LM. We further pinpoint that using {\ours} allows us to perform soft reasoning with simulated search, making it flexible for diverse reasoning benchmarks. We introduce a method to measure model reasoning faithfulness w.r.t. the problem formalization ingrained within {\ours}. Our results show that model performance is positively correlated with the faithfulness of the reasoning process. The systematic studies of the method show the benefits of using simulated search compared to natural language reasoning and external symbolic solvers. We further show that using {\ours} allows us to interpretably and rigorously detect hallucinations and sub-optimal and inconsistent reasoning patterns.

\section*{Limitations}
While FLARE offers significant improvements in faithfulness and interoperability, it depends on the quality of LLM-generated plans and code; errors or omissions in formalisation can propagate through the simulated search, potentially leading to incorrect or incomplete answers. The generation of formal code and simulated search traces depends heavily on the LLM's prompt-following ability. The simulation of code execution may not fully explore extremely large or open-ended problem spaces, and prompt sensitivity can affect search thoroughness.

\section*{Risks and Impact Statement}
FLARE advances the capabilities of large language models in logical reasoning and problem-solving, with potential positive impacts on applications requiring transparent and verifiable decision-making processes.
The ability of the method to formalise reasoning steps and detect inconsistencies could improve reliability in high-stakes domains like healthcare decision support, educational assessment, and automated planning systems.
However, this advancement also raises important considerations --- the improved reasoning capabilities could be misused to automate deceptive or manipulative argumentation, and the increased persuasiveness through logical formalisation could mask underlying biases or false premises.
Additionally, while FLARE improves transparency in reasoning, it may create a false sense of rigour in cases where the underlying logic is flawed but presented in a formally convincing manner.

\section*{Acknowledgments}
$\begin{array}{l}\includegraphics[width=1cm]{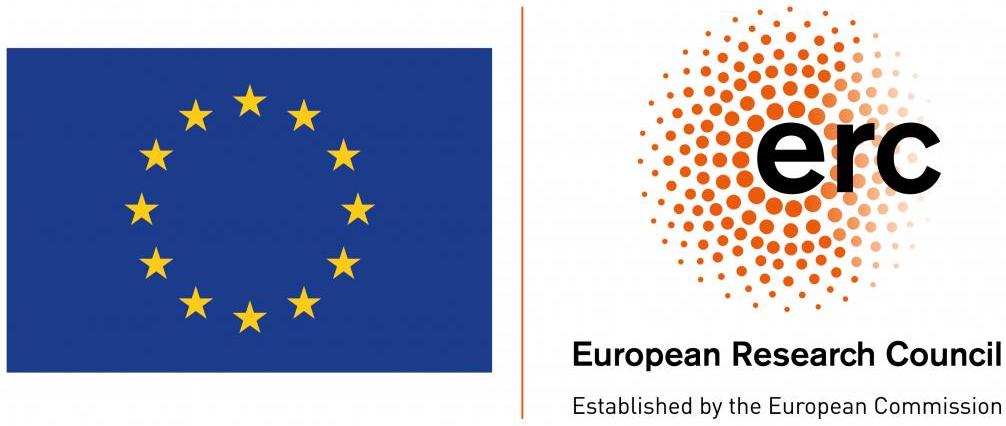} \end{array}$ 
Erik is partially funded by a DFF Sapere Aude research leader grant under grant agreement No 0171-00034B, as well as by an NEC PhD fellowship, and is supported by the Pioneer Centre for AI, DNRF grant number P1.
Pasquale was partially funded by ELIAI (The Edinburgh Laboratory for Integrated Artificial Intelligence), EPSRC (grant no.\ EP/W002876/1), an industry grant from Cisco, and a donation from Accenture LLP.
Isabelle's research is partially funded by the European Union (ERC, ExplainYourself, 101077481), and is supported by the Pioneer Centre for AI, DNRF grant number P1.
This work was supported by the Edinburgh International Data Facility (EIDF) and the Data-Driven Innovation Programme at the University of Edinburgh.
%



\bibliography{custom}
\clearpage
\appendix

\section{Appendix}
\begin{figure}[t!]
    \centering
    \includegraphics[clip=true,width=\columnwidth]{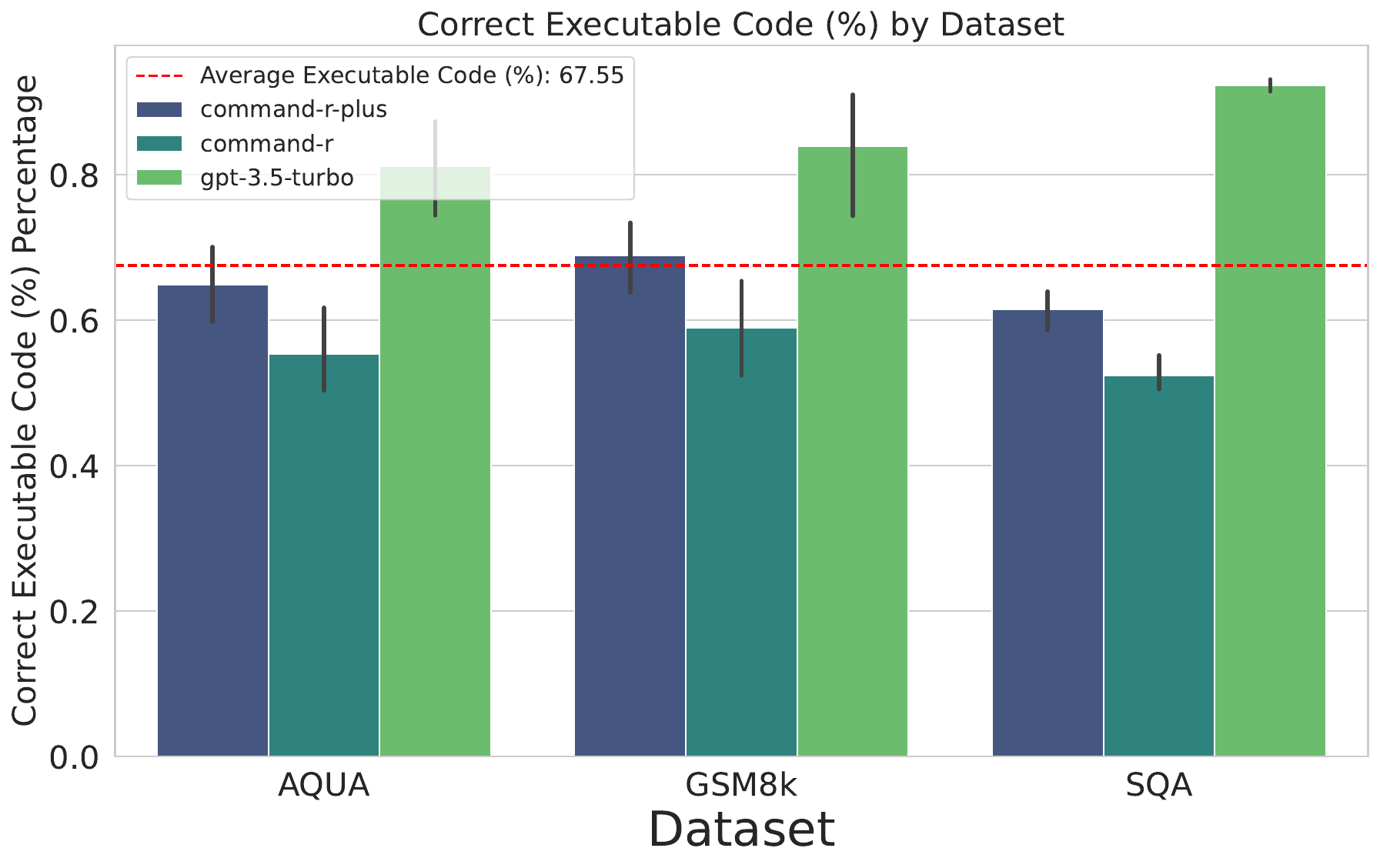}
    \includegraphics[clip=true,width=\columnwidth]{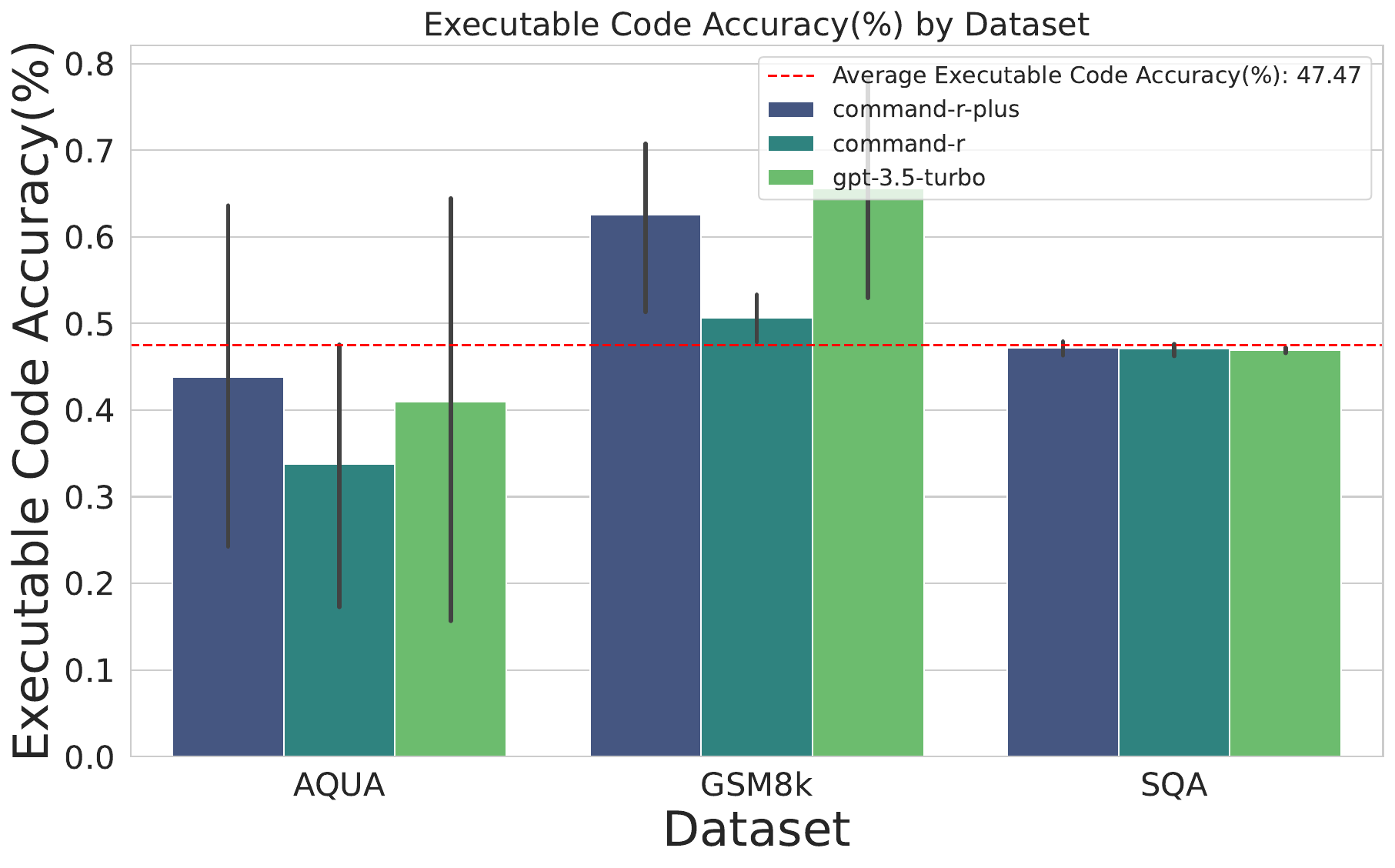}
    \caption{The figure shows the percentage of executable code per model (top) and the accuracy of the executable code when answering the queries (bottom).}
    \label{fig:code_perc}
\end{figure}

\begin{figure*}[t!]
    \centering
    \includegraphics[clip=true,width=0.9\textwidth]{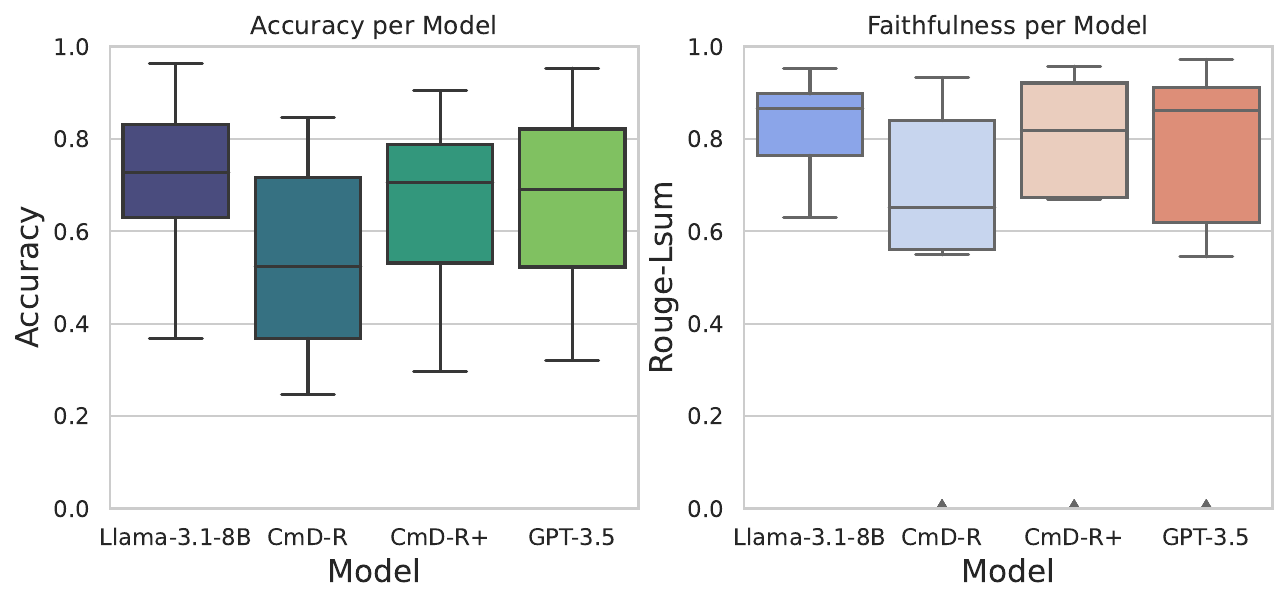}
    \caption{The effect of the model parameter scale from 8B to 100B+ on model accuracy (left) and faithfulness (right).}
    \label{fig:effect_of_scale}
\end{figure*}

\subsection{The effect of scale}
\label{sec:scale_effect}
\input{scale_effect}

We want to assess the impact of the number of parameters in the model on the overall performance and faithfulness. The results in \cref{fig:effect_of_scale} show no precise relation between model scale, performance and faithfulness. However, scaled models from the same family, i.e. CmDR (30B) and CmDR+ (100B), show improvements in reasoning faithfulness and model performance. We can also see in  \cref{tab:scale_effect} that as the model size increases, the average number of hops and the portion of hallucinations and unutilised knowledge decreases. This further confirms our prior assumptions that models with strong commonsense soft-reasoning capabilities can skip steps during the search while maintaining the knowledge and structure of the traversal strategy outlined in the code. 

\label{append:prompts}

\subsection{LLM Prompts}
We define straight-forward prompts for generating \emph{plan}, \emph{code} and \emph{search} simulation in {\ours}, which can be observed in \cref{tab:prompts}. 

\subsection{Dataset Statistics}

The datasets used in this study encompass a variety of domains, specifically targeting the performance of the models in interpreting Math Word Problems, multi-hop question answering, and relational inference. Table \ref{tab:data_stat} provides a detailed breakdown of each dataset, including the number of few-shot in-context samples (shots), the number of test samples, and representative examples from each dataset. The datasets provide a comprehensive basis for evaluating the models' abilities to handle complex tasks across different domains, facilitating an in-depth analysis of model performance under few-shot conditions.

\subsection{FLARE Pseudo-code}

Below, we present the pseudo-code for the execution of the \emph{plan}, \emph{code}, and \emph{search} procedures in {\ours}. The pseudo-code describes the modular pipeline in FLARE for tackling natural language queries with faithful simulated search. (i) \textbf{Plan Generation:} This stage creates a structured natural language outline of the reasoning process, breaking down the query into logical steps and analysis. The plan serves as the foundation for formalization into a logic-based representation. (ii) \textbf{Code Generation:} Based on the generated plan, a logic programming code (e.g., in Prolog) is synthesized. This code formalizes the query into a set of facts, relations, and goals, which collectively define the problem space for reasoning. (iii) \textbf{Search Simulation:} The generated code is utilized to simulate a search trace over the problem space. This includes iterative reasoning, backtracking when goals are unmet, and extracting emergent facts or relations during the process.
Each of these stages is implemented as a modular component. The generation from each of the stages feeds into the next, allowing seamless integration and incremental improvement in reasoning accuracy. A detailed pseudo-code is provided below in \cref{algo:flare}.

\input{data_stat}

\subsection{Benefits of Prolog}
\label{append:prolog}

Prolog is a symbolic logic-programming engine \citep{DBLP:conf/acm/Bowen79} used for heuristic search over Horn Clauses \citep{DBLP:journals/jlp/ChandraH85}. It is a declarative programming paradigm \citep{DBLP:conf/agp/Lloyd94}, meaning that the code is expressed as the logic of computation. In particular, this logic is formalised as a set of facts $\mathcal{F}$ and relations $\mathcal{R}$ forming our problem space, while the final goal $\mathcal{G}$ is a first-order logic combination of them. As a default, Prolog uses a depth-first search (DFS) strategy \citep{DBLP:conf/acm/Bowen79} for sub-goal decomposition and feasible traversal of the problem space that satisfies the goal $\mathcal{G}$. Such a traversal is referred to as the \emph{trace}. At each trace step, the program can either confirm or invalidate the sub-goal using the feasibility of fact and relation combinations, expand the search tree or retry satisfying a failed sub-goal with new combinations. An example of such a search can be observed in \cref{fig:flare}.
It is possible to complete an exhaustive search, exploring all possible paths that do or do not satisfy the goal. 
These characteristics are beneficial as we can explicitly access and segment the facts and relations that form the problem space and the search strategy used for query formalisation. As Prolog is declarative, it is sufficient to use a regexp heuristic for the segmentation, which is referred to as EXTRACT in \cref{eq:code} and \cref{eq:search}.
Furthermore, including exhaustive traversal traces in-context allows the LLM to simulate sub-goal decomposition, backtracking, intermediate goal invalidation, etc. We discuss this in more depth in the next paragraph.

\begin{algorithm*}[!t]
\label{algo:flare}
\caption{FLARE Methodology: Faithful Logic-Aided Reasoning and Exploration}
\begin{algorithmic}[1]
\Require Query $\mathcal{Q}$, Language Model $\mathcal{M}$
\Ensure Answer $\mathcal{A}$
\State \textbf{Initialization:} Load few-shot examples for plans ($\mathcal{E}_P$), code ($\mathcal{E}_C$), and search traces ($\mathcal{E}_S$)
\State \textbf{Input:} Natural language query $\mathcal{Q}$

\Procedure{Generate Plan}{}
    \State Prompt $\mathcal{M}$ with instructions $\mathcal{I}_P$ and examples $\mathcal{E}_P$ to generate a plan $\mathcal{P}$
    \State $\mathcal{P} \gets \arg\max \; p_{\mathcal{M}}(T_P|T_{P:<i}, \mathcal{E}_P, \mathcal{Q}, \mathcal{I}_P)$
\EndProcedure

\Procedure{Generate Code}{}
    \State Append examples $\mathcal{E}_C$ to $\mathcal{E}_P$
    \State Prompt $\mathcal{M}$ with instructions $\mathcal{I}_C$ to generate logic programming code $\mathcal{C}$
    \State $\mathcal{C} \gets \arg\max \; p_{\mathcal{M}}(T_C|T_{C:<i}, \mathcal{E}_C, \mathcal{Q}, \mathcal{P}, \mathcal{I}_C)$
    \State $(F_{\text{code}}, R_{\text{code}}, G_{\text{code}}) \gets \textsc{Extract}(\mathcal{C})$
\EndProcedure

\Procedure{Simulate Search}{}
    \State Append search trace examples $\mathcal{E}_S$ to $\mathcal{E}_C$
    \State Prompt $\mathcal{M}$ with instructions $\mathcal{I}_S$ to simulate a search trace $\mathcal{S}$
    \State $\mathcal{S} \gets \arg\max \; p_{\mathcal{M}}(T_S|T_{S:<i}, \mathcal{E}_S, \mathcal{Q}, \mathcal{P}, \mathcal{C}, \mathcal{I}_S)$
    \State $(F_{\text{search}}, R_{\text{search}}, \mathcal{A}_{\text{search}}) \gets \textsc{Extract}(\mathcal{S})$
    \While{Goal $\mathcal{G}_{\text{code}}$ is not satisfied}
        \State Explore next sub-goal in $\mathcal{S}$
        \If{Sub-goal fails}
            \State Backtrack to the previous state (Learned through in-context sampels)
        \EndIf
    \EndWhile
\EndProcedure

\Procedure{Final Answer Generation}{}
    \State Append correct answers from $\mathcal{A}_{\text{search}}$ to examples
    \State Prompt $\mathcal{M}$ with instructions $\mathcal{I}_A$ to finalize answer $\mathcal{A}$
    \State $\mathcal{A} \gets \arg\max \; p_{\mathcal{M}}(T_A|T_{A:<i}, \mathcal{E}_A, \mathcal{Q}, \mathcal{P}, \mathcal{C}, \mathcal{S}, \mathcal{I}_A)$
\EndProcedure

\State \Return $\mathcal{A}$
\end{algorithmic}
\end{algorithm*}

\input{prompts}
\input{complete}

\end{document}

%% file: results.tex
\begin{table*}[t!]
\adjustbox{width=\textwidth,center}{
\begin{tabular}{@{}lccccccccc@{}}
\toprule
\multicolumn{1}{c}{} &
  \multicolumn{5}{c}{\textbf{Math Word Problems}} &
  \multicolumn{3}{c}{\textbf{Multi-hop QA}} &
  \textbf{Relation} \\ \midrule
\multicolumn{1}{c|}{\textbf{Method}} &
  GSM8K &
  SVAMP &
  MultiArith &
  ASDiv &
  \multicolumn{1}{c|}{AQuA} &
  StrategyQA &
  Date &
  \multicolumn{1}{c|}{Sport} &
  CLUTRR \\ \midrule
\multicolumn{1}{l|}{$\text{Llama-3.1-8B}_\text{\ours}$} &
  \underline{72.7} &
  \textbf{86.0} &
  \textbf{96.3} &
  \textbf{83.1} &
  \multicolumn{1}{c|}{\cellcolor[HTML]{9AFF99}\textbf{62.9}} &
  \textbf{70.2} &
  \textbf{59.3} &
  \multicolumn{1}{c|}{\underline{76.6}} &
  {\underline{36.8}} \\
\multicolumn{1}{l|}{$\text{Llama-3.1-8B}_\text{F-CoT}$} &
  \textit{0} &
  \textit{0} &
  \textit{0} &
  \textit{0} &
  \multicolumn{1}{c|}{\textit{12.2}} &
  {\underline{53.2}} &
  \textit{0} &
  \multicolumn{1}{c|}{\textit{0}} &
  \textit{32} \\
\multicolumn{1}{l|}{$\text{Llama-3.1-8B}_\text{CoT}$} &
  \textbf{85.2} &
  82.4 &
  91.6 &
  79.1 &
  \multicolumn{1}{c|}{51.6} &
  \textit{43.5} &
  74.1 &
  \multicolumn{1}{c|}{\textbf{89.4}} &
  \textbf{45.7} \\
 \midrule
\multicolumn{1}{l|}{$\text{CmDR}_\text{\ours}$} &
  \textbf{52.4} &
  \textbf{74.0} &
  \textbf{84.5} &
  \textbf{72.2} &
  \multicolumn{1}{c|}{\textbf{43.7}} &
  \textbf{67.0} &
  \textbf{52.3} &
  \multicolumn{1}{c|}{\textbf{78.9}} &
  {\underline{29.1}} \\
\multicolumn{1}{l|}{$\text{CmDR}_\text{F-CoT}$} &
  \textit{0} &
  \textit{0} &
  \textit{0} &
  \textit{0} &
  \multicolumn{1}{c|}{\textit{0}} &
  {\underline{59.7}} &
  \textit{0} &
  \multicolumn{1}{c|}{\textit{0}} &
  \textit{8.6} \\
\multicolumn{1}{l|}{$\text{CmDR}_\text{CoT}$} &
  {\underline{46.5}} &
  {\underline{57.3}} &
  {\underline{83.1}} &
  {\underline{37.2}} &
  \multicolumn{1}{c|}{{\underline{28.3}}} &
  \textit{21.3} &
  {\underline{47.4}} &
  \multicolumn{1}{c|}{{\underline{55.2}}} &
  \textbf{29.5} \\ \midrule
\multicolumn{1}{l|}{$\text{CmDR+}_\text{\ours}$} &
  \textbf{71.4} &
  \cellcolor[HTML]{9AFF99}\textbf{83.5} &
  \textbf{90.4} &
  \textbf{81.3} &
  \multicolumn{1}{c|}{\textbf{55.9}} &
  \cellcolor[HTML]{9AFF99}\textbf{70.8} &
  {\underline{61.8}} &
  \multicolumn{1}{c|}{\textbf{77.7}} &
  \textbf{41.0} \\
\multicolumn{1}{l|}{$\text{CmDR+}_\text{F-CoT}$} &
  \textit{0} &
  \textit{0} &
  \textit{0} &
  \textit{0} &
  \multicolumn{1}{c|}{\textit{15.4}} &
  {\underline{57.6}} &
  \textit{0} &
  \multicolumn{1}{c|}{\textit{0}} &
  \textit{35.3} \\
\multicolumn{1}{l|}{$\text{CmDR+}_\text{CoT}$} &
  {\underline{48.7}} &
  {\underline{81.1}} &
  {\underline{86.6}} &
  {\underline{44.6}} &
  \multicolumn{1}{c|}{{\underline{44.1}}} &
  \textit{48.4} &
  \textbf{79.1} &
  \multicolumn{1}{c|}{{\underline{62.6}}} &
  {\underline{42.5}} \\ \midrule
\multicolumn{1}{l|}{$\text{GPT-3.5}_\text{\ours}$} &
  \cellcolor[HTML]{9AFF99}\textbf{82.1} &
  {\underline{82.7}} &
  \cellcolor[HTML]{9AFF99}\textbf{98.3} &
  \cellcolor[HTML]{9AFF99}\textbf{85.4} &
  \multicolumn{1}{c|}{{\underline{55.1}}} &
  \textbf{65.5} &
  \cellcolor[HTML]{9AFF99}\textbf{82.4} &
  \multicolumn{1}{c|}{{\underline{85.6}}} &
  \cellcolor[HTML]{9AFF99}\textbf{49.8} \\
\multicolumn{1}{l|}{$\text{GPT-3.5}_\text{F-CoT}$} &
  {\textit{75.8}} &
  \textbf{83.0} &
  \textit{95.3} &
  {\underline{81.7}} &
  \multicolumn{1}{c|}{\textit{53.5}} &
  \textit{51.5} &
  {\underline{73.5}} &
  \multicolumn{1}{c|}{\textit{52.3}} &
  {\underline{12.1}} \\
\multicolumn{1}{l|}{$\text{GPT-3.5}_\text{CoT}$} &
  \underline{79.8} &
  \textit{82.4} &
  {\underline{98.2}} &
  \textit{75.8} &
  \multicolumn{1}{c|}{\textbf{59.4}} &
  {\underline{51.7}} &
  \textit{69.9} &
  \multicolumn{1}{c|}{\cellcolor[HTML]{9AFF99}\textbf{95.8}} &
  {\underline{4.3}} \\ \bottomrule
\end{tabular}
}
\caption{The following table shows the performance of each of the tested models given a technique for reasoning. Each \textbf{bold}, \underline{underlined}, and \textit{italicised} element highlights the best, second best and worst technique per specific model. The overall best method per dataset is highlighted in \colorbox[HTML]{9AFF99}{green}.}
\label{tab:results_main}
\end{table*}

%% file: logiclm_compare.tex
\begin{table*}[t!]
\centering
\adjustbox{width=\textwidth}{
\begin{tabular}{@{}cccccc|ccccc@{}}
\toprule
\multirow{2}{*}{Dataset} & \multicolumn{5}{c|}{ChatGPT (gpt-3.5-turbo)} & \multicolumn{5}{c}{GPT-4 (gpt-4o)} \\ \cmidrule(lr){2-6} \cmidrule(l){7-11} 
                           & Standard & CoT   & Logic-LM       & \ours          & \ours$_{SR=2}$ & Standard & CoT    & Logic-LM       & \ours          & \ours$_{SR=2}$ \\
\midrule
PrOntoQA                  & 47.40    & 67.80 & 61.00          & 73.40          & \textbf{79.40}          & 77.40    & 98.79  & 83.20          & 98.87          & \textbf{99.24}       \\
LogicalDeduction          & 40.00    & 42.33 & \textbf{65.67} & 58.60          & 64.43                   & 71.33    & 75.25  & 87.63          & 88.00          & \textbf{90.33}       \\
AR-LSAT                   & 20.34    & 17.31 & 26.41          & 27.39          & \textbf{30.73}          & 33.33    & 35.06  & 43.04          & 39.82          & \textbf{45.02}       \\
\bottomrule
\end{tabular}
}
\caption{Comparison of results across datasets for ChatGPT (gpt-3.5-turbo) and GPT-4 (gpt-4o) using Standard, CoT, Logic-LM, FLARE, and $\text{FLARE}_{SR=2}$ approaches. \textit{SR=2} refers to a maximum of $2$ iterations of code self-refinement.}
\label{tab:logiclm_res}
\end{table*}

%% file: prompt_only.tex
\begin{table*}[t!]
\adjustbox{width=\textwidth}{
\begin{tabular}{@{}lcccccc@{}}
\toprule
\textbf{Method} &
  \multicolumn{1}{l}{$\text{CmDR}_\text{plan-only}$} &
  \multicolumn{1}{l}{$\text{CmDR}_\text{\ours}$} &
  \multicolumn{1}{l}{$\text{CmDR+}_\text{plan-only}$} &
  \multicolumn{1}{l}{$\text{CmDR+}_\text{\ours}$} &
  \multicolumn{1}{l}{$\text{GPT-3.5}_\text{plan-only}$} &
  \multicolumn{1}{l}{$\text{GPT-3.5}_\text{\ours}$} \\ \midrule
GSM8K      & 24.7          & \textbf{52.4} & 40.7          & \textbf{71.4} & 36.1          & \textbf{68.1} \\
AQuA       & 35.0          & \textbf{43.7} & 55.1          & \textbf{55.9} & 54.3          & \textbf{55.1} \\
StrategyQA & 65.5          & \textbf{67.0} & \textbf{75.7} & 70.8          & 62.3          & \textbf{65.5} \\ \bottomrule
\end{tabular}
}
\caption{The table shows the accuracy of an LLM with {\ours} compared to prompting for a final answer directly after generating (plan-only) a plan $\mathcal{P}$.}
\label{tab:plan_only}
\end{table*}

%% file: reasoning_stats_basic.tex
\begin{table}[t!]
\centering
\adjustbox{max width=\linewidth}{
\begin{tabular}{@{}lccccc@{}}
\toprule
\textbf{Model} & \textbf{\#Paths} & \textbf{\#Hops/p} & \textbf{\#Fails/p} & \textbf{TotHops} & \textbf{TotFails} \\ 
\midrule
\multicolumn{6}{c}{\textbf{Incorrect Answers}} \\
$\text{Llama-3.1-8B}_\text{\ours}$ & 1.55 & 11.12 & \cellcolor[HTML]{CBCEFB}1.52 & 15.09 & \cellcolor[HTML]{CBCEFB}2.26 \\
$\text{CmDR}_\text{\ours}$         & 1.51 & 6.55  & \cellcolor[HTML]{CBCEFB}0.68 & 10.56 & \cellcolor[HTML]{CBCEFB}1.39 \\
$\text{CmDR+}_\text{\ours}$        & 0.92 & 7.52  & \cellcolor[HTML]{CBCEFB}1.13 & 8.57  & \cellcolor[HTML]{CBCEFB}1.32 \\
GPT-3.5                             & 0.68 & 5.22  & \cellcolor[HTML]{CBCEFB}0.71 & 5.32  & \cellcolor[HTML]{CBCEFB}0.74 \\ 
\midrule
\multicolumn{6}{c}{\textbf{Correct Answers}} \\
\midrule
$\text{Llama-3.1-8B}_\text{\ours}$ & 1.43 & 9.12  & 0.62                         & 12.36 & 0.96                         \\
$\text{CmDR}_\text{\ours}$         & 1.19 & 7.10  & 0.42                         & 11.29 & 0.66                         \\
$\text{CmDR+}_\text{\ours}$        & 0.97 & 7.19  & 0.42                         & 8.22  & 0.61                         \\
$\text{GPT-3.5}_\text{\ours}$      & 0.82 & 5.65  & 0.26                         & 5.69  & 0.27                         \\ 
\bottomrule
\end{tabular}
}
\caption{%
\textbf{\#Paths}: Avg.\ number of reasoning paths tried by the model. 
\textbf{\#Hops/p}: Avg.\ number of hops per path. 
\textbf{\#Fails/p}: Avg.\ number of fails (unsuccessful hops) per path. 
\textbf{TotHops}: Avg.\ total hops (summed across all paths).
\textbf{TotFails}: Avg.\ total fails (summed across all paths).
The \cellcolor[HTML]{CBCEFB}purple cells show that incorrect reasoning paths often have fewer failed search paths.
}
\label{tab:reasoning_stats}
\end{table}

%% file: reasoning_stats_deep.tex
\begin{table}[!t]
\adjustbox{width=\columnwidth}{
\begin{tabular}{@{}lccc@{}}
\toprule
\textbf{Model} & \multicolumn{1}{r}{UEF (\%) in Search} & \multicolumn{1}{l}{OR (\%)} & \multicolumn{1}{r}{UR (\%) in code} \\ \midrule
\multicolumn{4}{c}{\textbf{Correct Answers}}   \\ \midrule
$\text{Llama-3.1-8B}_\text{\ours}$   & 74.14   & 43.65  & 5.73  \\
$\text{CmDR}_\text{\ours}$               & 59.06   & 35.96  & 4.02  \\
$\text{CmDR+}_\text{\ours}$              & 64.30   & 34.47  & 4.54  \\
$\text{GPT-3.5}_\text{\ours}$             & 64.46   & 37.55  & 1.90  \\ \midrule
\textbf{Avg. (Correct)}               & 65.49   & 37.91  & 4.05  \\ \midrule
\multicolumn{4}{c}{\textbf{Incorrect Answers}} \\ \midrule
$\text{Llama-3.1-8B}_\text{\ours}$   & 54.69   & 35.04  & 9.28  \\
$\text{CmDR}_\text{\ours}$               & 54.50   & 32.76  & 6.23  \\
$\text{CmDR+}_\text{\ours}$              & 44.12   & 24.98  & 8.22  \\
$\text{GPT-3.5}_\text{\ours}$             & 36.02   & 24.44  & 6.94  \\ \midrule
\textbf{Avg. (Incorrect)}             & 47.33   & 29.31  & 7.67  \\ \midrule
\textbf{$\Delta$}     & 18.16   &  8.60  & -3.62 \\ \bottomrule
\end{tabular}
}
\caption{The table shows how the percentage of unique emergent facts (UEF) in search, overlapping relations (OR) between code and search, and unused relations (UR) in code impact answer correctness.}
\label{tab:reasoning_stats_deep}
\end{table}

%% file: scale_effect.tex

\begin{table}[t!]
    \resizebox{\columnwidth}{!}{%
    \begin{tabular}{@{}lccc@{}}
        \toprule
        \textbf{Model} &
        \textbf{Avg. hops per path} &
        \textbf{Hal.\,(\%)} &
        \textbf{UK.\,(\%)} \\ \midrule
        Llama-3.1-8B & 9.4 & 63.3 & 62.9 \\
        CmDR         & 6.7 & 54.7 & 56.9 \\
        CmDR+        & 7.2 & 54.3 & 56.3 \\
        GPT-3.5      & 5.5 & 49.3 & 52.1 \\ \bottomrule
    \end{tabular}}
    \caption{Changes in simulated search statistics when using {\ours} across model scales (8B to 100B+). \emph{Hallucinations} (Hal.) are facts/predicates used only in the trace, while \emph{unutilised knowledge} (UK.) denotes facts/relations appearing only in the code.}
    \label{tab:scale_effect}
\end{table}

%% file: data_stat.tex
\begin{table*}[t!]
\adjustbox{width=\textwidth}{
\begin{tabular}{@{}ccccc@{}}
\toprule
Domain &
  Dataset &
  Shots &
  Test Samples &
  Example \\ \midrule
\multirow{4}{*}{\begin{tabular}[c]{@{}c@{}}Math \\ Word \\ Problems\end{tabular}} &
  GSM8K &
  8 &
  1,319 &
  \begin{tabular}[c]{@{}c@{}}Q: A robe takes 2 bolts of blue fiber and half that much white fiber.  \\ How many bolts in total does it take?\\ A: 3\end{tabular} \\
 &
  SVAMP &
  8 &
  1,000 &
  Q: Dan had \$3 left with him after he bought a candy bar. If he had \$4 at the start, how much did the candy bar cost?A: 1 \\
 &
  MultiArith &
  8 &
  600 &
  \begin{tabular}[c]{@{}c@{}}Q: A pet store had 13 siamese cats and 5 house cats. During a sale they sold 10 cats. \\ How many cats do they have left? \\ A: 8\end{tabular} \\
 &
  ASDiv &
  8 &
  2,096 &
  \begin{tabular}[c]{@{}c@{}}Q: Adam has five more apples than Jackie. Jackie has nine apples. How many apples does Adam have?\\ \\ A: 14\end{tabular} \\
\multicolumn{1}{l}{} &
  AQuA &
  8 &
  254 &
  \begin{tabular}[c]{@{}c@{}}Q: A man walks at 5 kmph for 6 hrs and at 4 kmph for 12 hrs. His average speed is\\ Answer option: A)4 1/3 km/h, B)7 2/3 km/h, C)9 ½ km/h, D)8 km/h, E)81 km/h\\ A: A\end{tabular} \\ \midrule
\begin{tabular}[c]{@{}c@{}}Multi- \\ hop \\ QA\end{tabular} &
  StrategyQA &
  6 &
  2,290 &
  \begin{tabular}[c]{@{}c@{}}Q: Did Aristotle use a laptop? \\ A: False\end{tabular} \\
 &
  \begin{tabular}[c]{@{}c@{}}Date \\ Understanding\end{tabular} &
  10 &
  359 &
  \begin{tabular}[c]{@{}c@{}}Q: Yesterday was April 30, 2021. What is the date tomorrow in MM/DD/YYYY? \\ A: "05/02/2021"\end{tabular} \\
 &
  \begin{tabular}[c]{@{}c@{}}Sports \\ Understanding\end{tabular} &
  10 &
  977 &
  \begin{tabular}[c]{@{}c@{}}Q: Is the following sentence plausible? Lionel Messi was called for icing? \\ A: False\end{tabular} \\ \midrule
\begin{tabular}[c]{@{}c@{}}Relational \\ Inference\end{tabular} &
  CLUTRR &
  8 &
  1,042 &
  \begin{tabular}[c]{@{}c@{}}Q: [Carlos] is [Clarence]'s brother. [Carlos] and his sister, [Annie], went shopping. \\ asked her mom [Valerie] if she wanted anything, but [Valerie] said no. \\ How is [Valerie] related to [Clarence]? \\ A: "mother"\end{tabular} \\ \bottomrule
\end{tabular}
}
\caption{The statistics and examples of the datasets used in benchmarking. Shots refers to the number of few-shot in-context samples used during benchmarking.}
\label{tab:data_stat}
\end{table*}

%% file: prompts.tex
\begin{table*}[t!]
\label{tab:prompts}
    \centering
    \adjustbox{width=\textwidth}{
    \begin{tabular}{@{}p{3cm}p{5cm}p{5cm}@{}}
        \toprule
        \textbf{Task} & \textbf{Prompt} & \textbf{Description} \\
        \midrule
        \textbf{Plan Generation} & 
        \begin{minipage}[t]{5cm}
        \vspace{0pt}
        Generate an explanation and analysis, and plan to generate a prompt for writing a swi-prolog code for the last task. The 3 sections should be exactly outlined. Your plan should show enough intermediate reasoning steps towards the answer. Construct the plan as much as you can and describe the logic specifically. When constructing the plan for the code prompt, actively use swi prolog search capabilities.
        \end{minipage}
        & 
        \begin{minipage}[t]{5cm}
        \vspace{0pt}
        Detailed instructions for generating an outline and plan, with an emphasis on reasoning steps and using Prolog's search capabilities.
        \end{minipage}
        \\
        \midrule
        \textbf{Code Generation} & 
        \begin{minipage}[t]{5cm}
        \vspace{0pt}
        Write a Prolog code to solve using the plan. If there are unknown or stochastic atoms or predicates, fill in the values for them as a logical assumption and add a comment in the same line Assumed atom/predicate". Do not use write and read commands within the code. The code should be very detailed and utilize swi prolog capabilities to the fullest. To run the program, at the end create a predicate named "query" that returns the correct numerical answer. The last line of the program should be the commented-out driver predicate "query". Write only the code.
        \end{minipage}
        & 
        \begin{minipage}[t]{5cm}
        \vspace{0pt}
        Instructions for generating a Prolog code based on the plan with assumptions for unknown atoms. Emphasizes code details and a final ``query'' predicate.
        \end{minipage}
        \\
        \midrule
        \textbf{Simulated Search} & 
        \begin{minipage}[t]{5cm}
        \vspace{0pt}
        Ignoring the read commands, explicitly write out the search paths that are explored by the code:
        \#\#\#\#
        Here are the paths [Starting Search Simulation]: 
        \#\#\#\# 
        [Path 1]:
        \end{minipage}
        & 
        \begin{minipage}[t]{5cm}
        \vspace{0pt}
        A task to simulate and display the search paths that the Prolog code would follow during execution.
        \end{minipage}
        \\
        \midrule
        \textbf{Final Answer} & 
        \begin{minipage}[t]{5cm}
        \vspace{0pt}
        Given the plan, the code and the explored search paths answer the question above. Answer with the correct numerical answer.
        \#\#\#\#\# Here is the answer:
        \end{minipage}
        & 
        \begin{minipage}[t]{5cm}
        \vspace{0pt}
        Final prompt asking for the correct numerical answer based on the previous steps.
        \end{minipage}
        \\
        \bottomrule
    \end{tabular}
    }
    \caption{Table of Prompts for Plan, Code, Simulated Search, and Final Answer generation for GSM8K \citep{DBLP:journals/corr/abs-2110-14168}.}
\end{table*}

%% file: complete.tex
\begin{table*}[!t]
    \centering
    \begin{tabular}{|p{2cm}|p{13cm}|}
        \hline
        \textbf{Section} & \textbf{Content} \\ \hline

        Question & "Can Reiki be stored in a bottle?" \\ \hline

        Explanation & Reiki is a form of alternative therapy that involves the channeling of energy to promote healing and relaxation. It is often administered through the practitioner's hands, with the energy flowing directly to the recipient. The question raises the possibility of storing Reiki energy in a bottle, which may be a common query due to the abstract nature of energy healing. \\ \hline

        Analysis & To address the question of whether Reiki can be stored in a bottle, it is important to consider the nature of Reiki energy, the principles of energy healing, and the practical implications of containing and preserving such energy. \\ \hline

        Plan & \begin{itemize}
            \item Understand Reiki Energy: ...
            \item Explore Energy Storage: ...
            \item Consider Symbolism and Intention: ...
            \item Evaluate Practicality: ...
            \item Reflect on Tradition: ...
        \end{itemize} \\ \hline

        Code & \begin{verbatim}
        % Define the nature of Reiki energy
        reiki_energy(spiritual_energy).
        reiki_energy(channeling_through_touch).
        % Define the feasibility of storing energy in a bottle
        store_energy_in_bottle(feasible) :- 
            reiki_energy(spiritual_energy),
            reiki_energy(channeling_through_touch).
        % Storing Reiki energy is not a common practice
        not_common_practice(store_reiki_in_bottle).
        % Query to determine if Reiki can be stored in a bottle
        query :-
            store_energy_in_bottle(feasible),
            not_common_practice(store_reiki_in_bottle).
        % :- query.
        \end{verbatim} \\ \hline

        Search Paths & 
            \begin{itemize}
              \setlength\itemsep{0.05em}

                \item 0: Start of execution: Beginning Search
                \item 1: Call: query
                \item 2: Call: store\_energy\_in\_bottle(feasible)
                \item 3: Call: reiki\_energy(spiritual\_energy)
                \item 4: Call: reiki\_energy(channeling\_through\_touch)
                \item 5: Call: not\_common\_practice(store\_reiki\_in\_bottle)
                \item 6: Call: feasible=no
                \item 7: Fail: feasible=no | {'Result': 'Search Failed'}
                \item 8: Fail: query | {'Result': 'Search Failed'}
        \end{itemize} \\ \hline

        Answer & "No, Reiki energy cannot be stored in a bottle based on the logical evaluation of its abstract, non-physical nature and traditional practices of Reiki." \\ \hline
    \end{tabular}
    \caption{Complete example of \ours}
    \label{tab:flare_ex_complete}
\end{table*}